# Storing and Indexing Plan Derivations through Explanation-based Analysis of Retrieval Failures

**Laurie H. Ihrig**                                                    IHRIG@ASU.EDU
**Subbarao Kambhampati**                                              RAO@ASU.EDU
*Department of Computer Science and Engineering*
*Arizona State University*
*Tempe, AZ 85287-5406*

## Abstract

Case-Based Planning (CBP) provides a way of scaling up domain-independent planning to solve large problems in complex domains. It replaces the detailed and lengthy search for a solution with the retrieval and adaptation of previous planning experiences. In general, CBP has been demonstrated to improve performance over generative (from-scratch) planning. However, the performance improvements it provides are dependent on adequate judgements as to problem similarity. In particular, although CBP may substantially reduce planning effort overall, it is subject to a mis-retrieval problem. The success of CBP depends on these retrieval errors being relatively rare. This paper describes the design and implementation of a replay framework for the case-based planner DERSNLP+EBL. DER-SNLP+EBL extends current CBP methodology by incorporating explanation-based learning techniques that allow it to explain and learn from the retrieval failures it encounters. These techniques are used to refine judgements about case similarity in response to feedback when a wrong decision has been made. The same failure analysis is used in building the case library, through the addition of repairing cases. Large problems are split and stored as single goal subproblems. Multi-goal problems are stored only when these smaller cases fail to be merged into a full solution. An empirical evaluation of this approach demonstrates the advantage of learning from experienced retrieval failure.

## 1. Introduction

*Case-Based Planning* improves the efficiency of plan generation by taking advantage of previous problem-solving experiences. It has been shown to be an effective method for scaling up domain-independent planning to solve large problems in complex domains (Kambhampati & Hendler, 1992; Veloso, 1994). CBP involves storing information about the particular planning episodes in which problems are successfully solved. This information may include the goals that were achieved, the world state conditions which were found to be relevant to their achievement, the final plan and the decisions that were taken in arriving at this plan. Whenever a new problem is encountered, a judgment is made about its similarity to these previous experiences. Similar cases are then reused and extended in the search for a solution to the new problem. For example, the previous plan may be transformed into a *skeletal plan* which is then further refined into a new solution (Friedland & Iwasaki, 1985; Kambhampati & Hendler, 1992; Hanks & Weld, 1995). Multiple cases, each corresponding to a small subproblem, may be combined and extended in solving a single larger problem (Redmond, 1990; Ram & Francis, 1996). Alternatively, plan derivations may be *replayed* to provide guidance to a new search process (Veloso, 1994; Ihrig & Kambhampati, 1994a). CBP improves problem-solving in that problems are solved in less time in comparison to generative planning.

One of the most challenging tasks in CBP is in determining which cases to store and how to match these cases to a new problem-solving context. In a complex domain, it is unlikely that the same problem will be seen more than once. Moreover, if every problem solved is stored, the library will be large and the cost associated with retrieval may overshadow any gains that it provides (Koehler, 1994; Francis & Ram, 1995). Ultimately, we would like to retain in the library a minimum number of cases such that all new problems are solved through the efficient retrieval and adaptation of the cases that are stored (Smyth & Keane, 1995). However, in complex domains, the planner's theory of problem similarity is incomplete (Barletta & Mark, 1988). It does not have information about all of the relevant features of a new situation which determine if a stored case will be applicable. Sometimes a new problem will contain extra goals and/or changed initial state conditions. These changes may mean that a solution cannot be found which is consistent with the earlier planning decisions made in a stored episode. If the planner cannot predict ahead of time that these previous choices are wrong for the current situation, it will experience a retrieval error.

In this paper, we introduce DERSNLP+EBL (DERivational Systematic NonLinear Planner + Explanation-Based Learning), a CBP system which like PRIAR (Kambhampati & Hendler, 1992) and SPA (Hanks & Weld, 1995) is based on a sound and complete domain-independent planner. DERSNLP+EBL deals with the mis-retrieval problem by allowing the planner to learn from its planning failures so that it may anticipate future errors. Failure explanations are automatically generated from the search process which is used in extending the case to the new problem-solving situation. These are used in building the case library through the addition of repairing cases.

Although earlier systems such as CHEF (Hammond, 1990) have exploited EBL techniques, their use is restricted to reasoning about the correctness of the plans generated by the case-based planner. In contrast, DERSNLP+EBL starts with a sound and complete plan synthesis strategy. The emphasis here is on improving the performance of the base-level planner through the guidance of the retrieved cases. This guidance is considered to succeed if it leads the planner down a search path leading to a solution to the new problem. A retrieval error occurs when the planner is directed down a wrong path in its search for a solution, that is, a path that does not lead to a solution. DERSNLP+EBL extends current CBP methodology through EBL techniques which are employed in the automatic generation of reasons for a retrieval failure. Analytical failures that occur in the leaf nodes of the search tree are explained in terms of subsets of conflicting plan constraints. These leaf node failure explanations are regressed up the failing search paths to form a reason for the retrieval failure.

DERSNLP+EBL builds and indexes the case library based on this failure analysis. The failure reason is used in the construction of a new repairing case. For example, if a retrieved case fails due to the presence of an extra interacting goal which is not covered in the retrieved episodes, an explanation for the failure is formed which identifies the subset of new input goals which are *negatively interacting*. The failure reason is used to construct a new case which solves those goals alone. The same failure analysis is also employed in refining the indexing in the case library to both censor the retrieval of the failing case whenever the interacting goals are present again, and to direct the retriever to a new repairing case which avoids the failure.

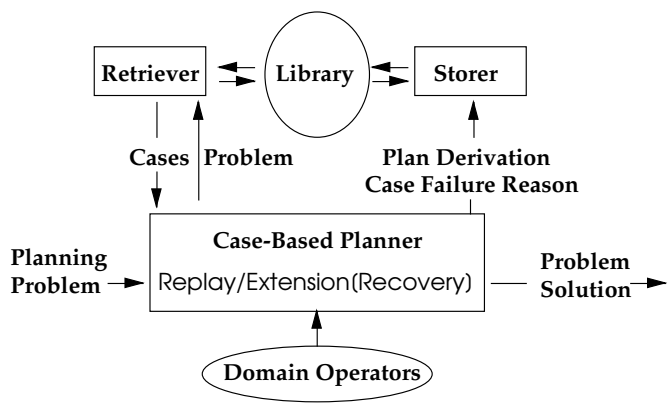

Figure 1: A schematic diagram illustrating the approach of DERSNLP+EBL.

DERSNLP+EBL's failure-based storage strategy limits the size of the case library. Library size is reduced by splitting problems into single goal subproblems, and storing these separately. Large problems are then solved through the retrieval and adaptation of multiple instances of these smaller cases. Multi-goal problems are stored only when the retrieved cases fail to be merged and extended into a full solution. We will describe empirical studies which demonstrate substantial improvements in performance with this novel approach to multi-case adaptation.

The remainder of the paper is organized as follows: Section 2 describes how DER-SNLP+EBL learns from case failure to improve its case retrieval. It also reports some preliminary experiments testing this learning component. Section 3 provides efficient techniques used to store, retrieve and adapt multiple cases. It describes experiments to test DERSNLP+EBL's method of plan merging. Section 4 describes an evaluation of the full DER-SNLP+EBL system when solving large problems drawn from a complex domain. Section 5 relates our work to previous case-based planners, including CHEF and PRODIGY/ANALOGY. Section 6 provides a summary.

## 2. Learning from Case Failure

As stated earlier, DERSNLP+EBL is based on a complete and correct domain-independent planning strategy. Like PRIAR (Kambhampati & Hendler, 1992) and SPA (Hanks & Weld, 1995), it is implemented on a partial-order planner. In this aspect it differs from state-space systems such as PRODIGY/ANALOGY (Veloso & Carbonell, 1993a; Veloso, 1994) and PARIS (Bergmann & Wilke, 1995). Like PRODIGY/ANALOGY, it employs the case adaptation strategy, *derivational replay* which stores planning experience in the form of successful plan derivations. Previous decisions made in earlier planning episodes become instructions to guide the search process in solving the new problem. Derivational replay includes all of the following elements, as illustrated in Figure 1 (Veloso, 1994; Veloso & Carbonell, 1993a): a facility within the underlying planner to generate a trace of the derivation of a plan, the indexing and storage of the derivation trace in a library of previous cases, the retrieval of multiple cases in preparation for solving a new problem, and finally, a replay mechanism by which the planner employs the retrieved plan derivations as a sequence of instructions to guide the new search process.

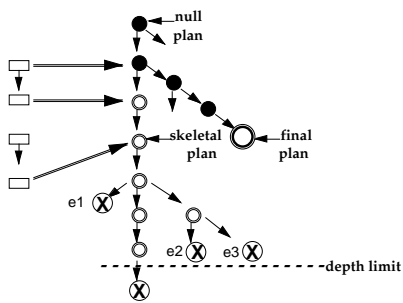

Figure 2: Multiple derivation traces (each a sequence of decisions shown in the figure as rectangles) are used to guide the new search process. In the figure, a solution could be reached only by backtracking over the skeletal plan, which now lies outside the new plan derivation (shown as filled circles).

DERSNLP+EBL's methodology depends on aspects that it has in common with MOLGEN (Friedland & Iwasaki, 1985) and PRIAR (Kambhampati & Hendler, 1992). It requires an *eager* case adaptation strategy, so that a skeletal plan will be constructed which contains all of the constraints added on the advice of the retrieved cases, and only those constraints. This is to separate the failure resulting from the previous guidance from any subsequent planning effort. In eager case adaptation, the planning decisions which are encapsulated in the retrieved cases are greedily adopted before these decisions are extended to solve any extra goals not covered. Multiple retrieved plan derivations are replayed in sequence to produce the skeletal plan which then contains all of the recommended plan constraints. The planner returns to from-scratch planning only *after* all of the previous decisions in the retrieved cases have been visited. The skeletal plan is then further refined to achieve any goals left open. Previous work has demonstrated the effectiveness of this approach to plan-space replay as well as its advantage over state-space replay (Ihrig & Kambhampati, 1994a, 1994b).

Eager case adaptation can also be described as *extension-first*. The skeletal plan is first extended in the search for a solution, and, only after this extension fails, is the plan backtracked over, discarding the plan constraints which were added on the advice of previous episodes. The general approach to case adaptation therefore involves three distinct phases: case replay, case extension, and, if extension fails, recovery. During the search process employed in extending the skeletal plan, the planner constructs an explanation for the plan's failure which becomes a reason for the case retrieval failure. Explanations are formed for the analytical failures that occur in the leaf nodes directly under the skeletal plan (See Figure 2). An analytical failure is explained by a set of inconsistent plan constraints. These failure explanations are immediately regressed up the search paths as they are encountered. The regressed explanations are collected at the root of the tree to form a reason for the retrieval error. DERSNLP+EBL detects that a retrieval error has occurred when all ways of refining the skeletal plan have been tried, and the planner is forced to backtrack over this plan. At this point the failure reason is fully constructed. Performing skeletal plan extension as a separate process prior to recovery allows the planner to identify the retrieval error in terms of the failure of the skeletal plan, and to construct a reason for the failure. This reason is then communicated to the Storer to be used in augmenting the library with a new repairing case.

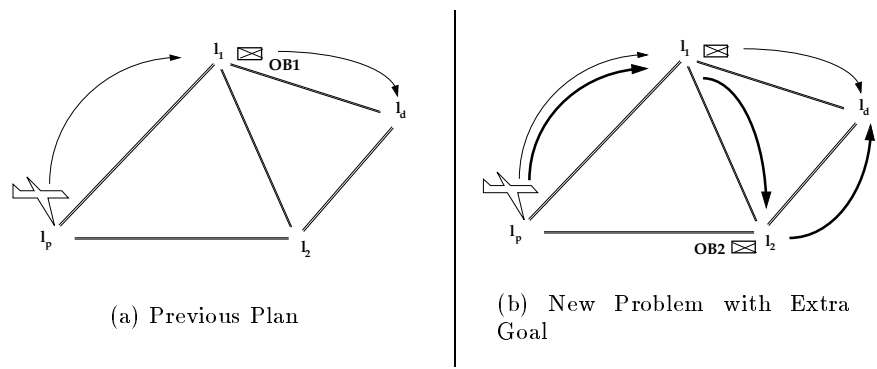

(a) Previous Plan

(b) New Problem with Extra Goal

Figure 3: (a) A plan to accomplish the transport of a single package, OB1, to the destination airport $l_d$. (b) A new problem contains an extra goal which involves the additional transport to $l_d$ of a second package, OB2.

Consider a simple example illustrated in Figure 3 which is taken from the Logistics Transportation domain shown in Figure 4. The goal is to have package OB1 located at the destination location $l_d$. The package is initially at location $l_1$. There is a plane located at $l_p$ which can be used to transport the package. Figure 3a illustrates a previous plan which contains steps that determine the plane's route to the destination airport as well as steps which accomplish the loading of the package at the right place along this route. Eagerly replaying these earlier step addition decisions for a new problem in which there is an extra package to transport produces a skeletal plan which may readily be extended to include the loading and unloading of the extra package as long as this package lies along the same route. However, if the new package is off the old route, the planner may not be able to solve the extra goal without backtracking over some of its previous step addition decisions. (See Figure 3b).

A case failure reason is shown in Figure 5. It gives the conditions under which a future replay of the case will again result in failure. These conditions refer to the presence in the new problem of a set, $\mathcal{C}$, of *negatively interacting* goals, as well as some initial state conditions, contained in $\mathcal{E}$. A summary of the information content of the failure reason is: *There is an extra package to transport to the same destination location, and that package is not at the destination location, is not inside the plane, and is not located on the plane's route.*

| action<br>precond<br><br>add<br>delete | (LOAD-TRUCK ?O ?T ?L)<br>(AT-OB ?O ?L)<br>(AT-TR ?T ?L)<br>(INSIDE-TR ?O ?T)<br>(AT-OB ?O ?L) | action<br>precond<br><br>add<br>delete | (LOAD-PLANE ?O ?P ?L)<br>(AT-OB ?O ?L)<br>(AT-PL ?P ?L)<br>(INSIDE-PL ?O ?P)<br>(AT-OB ?O ?L) |
| action<br>precond<br><br>add<br>delete | (UNLOAD-TRUCK ?O ?T ?L)<br>(INSIDE-TR ?O ?T)<br>(AT-TR ?T ?L)<br>(AT-OB ?O ?L)<br>(INSIDE-TR ?O ?T) | action<br>precond<br><br>add<br>delete | (UNLOAD-PLANE ?O ?P ?Li)<br>(INSIDE-PL ?O ?A)<br>(AT-PL ?P ?Li)<br>(AT-OB ?O ?Li)<br>(INSIDE-PL ?O ?A) |
| action<br>precond<br><br>add<br>delete<br>equals | (DRIVE-TRUCK ?T ?Li ?Lg)<br>(AT-TR ?T ?Li)<br>(SAME-CITY ?Li ?Lg)<br>(AT-TR ?T ?Lg)<br>(AT-TR ?T ?Li)<br>(NOT (?Li ?Lg)) | action<br>precond<br><br>add<br>delete<br>equals | (FLY-PLANE ?P ?Li ?Lg)<br>(IS-A AIRPORT ?Lg)<br>(AT-PL ?P ?Li))<br>(AT-PL ?P ?Lg)<br>(AT-PL ?P ?Li)<br>(NOT (?Li ?Lg)) |

Figure 4: The specification of the Logistics Transportation Domain adapted for our experiments

Subsequent to backtracking over the skeletal plan, the planner continues its search, and will go on to find a solution to the full problem if one exists. This new solution achieves all of the negatively interacting goals identified in the failure reason. Moreover, since these goals represent a subset of the problem goals, the new derivation may be used to construct a case covering these goals alone. DERSNLP+EBL stores the new case directly beneath the failing case so as to censor its retrieval. This is to ensure that whenever the failure reason holds (for example, whenever there is an extra package which is off the plane's route), the retriever is directed away from the failing case and toward the case that repairs the failure.

We are now in a position to describe in more detail DERSNLP+EBL's eager derivation replay strategy, as well as how it learns the reasons underlying a case failure.

## 2.1 Eager Derivation Replay

The derivation trace contains a sequence of instructions representing the choices that lie along the derivation path leading from the root of the search tree to the final plan in the leaf node. The trace is fitted to the context of the new search process by validating each choice in the new context, and replaying the decision if valid. In order to understand the validation process, we must first describe the decision steps that the planner takes in arriving at a solution to a planning problem. A planning problem is a 3-tuple $\langle I, G, \mathcal{A} \rangle$, where $I$ is a complete description of the initial state, $G$ is the description of the goal state, and $\mathcal{A}$ is the set of operators in STRIPS representation (Fikes & Nilsson, 1971). A ground operator sequence is said to be a solution for a planning problem if it can be executed from the initial state, and the resulting state of the world satisfies the goal.

DERSNLP+EBL is a refinement planner that solves a planning problem by navigating a space of potential solutions, each represented as a partly constructed plan[1]. Syntactically, a

---

1. For a more formal development of the refinement search semantics of partial plans, we refer the reader to the work of Kambhampati, Knoblock, and Yang (1995).

Case Failure Explanation:

$\mathcal{C} = \{$ ⟨(AT-OB OB1 $l_d$), $t_G$⟩, ⟨(AT-OB OB2 $l_d$), $t_G$⟩ $\}$
$\mathcal{E} = \{$ ⟨$t_I$, (¬AT-OB OB2 $l_d$)⟩, ⟨$t_I$, (¬INSIDE-PL OB2 ?PL )⟩,
⟨$t_I$, (¬AT-OB OB2 $l_1$)⟩, ⟨$t_I$, (¬AT-OB OB2 $l_p$)⟩ $\}$

Figure 5: An Example of a case failure reason

plan in this space $\mathcal{P}$ can be seen as a set of constraints (see below). Semantically, a partial plan is a shorthand notation for the set of ground operator sequences that are consistent with its constraints. The latter is called the *candidate set* of the partial plan, and is denoted by $\langle\!\langle \mathcal{P} \rangle\!\rangle$. In particular, a partial plan is represented as a 6-tuple, $\langle \mathcal{S}, \mathcal{O}, \mathcal{B}, \mathcal{L}, \mathcal{E}, \mathcal{C} \rangle$, where

1. $\mathcal{S}$ is the set of actions (step names) in the plan, each of which is mapped onto an operator in the domain theory. $\mathcal{S}$ contains two dummy steps: $t_I$ whose effects are the initial state conditions, and $t_G$ whose preconditions are the input goals, $G$.

2. $\mathcal{B}$ is a set of codesignation (binding) and non-codesignation (prohibited binding) constraints on the variables appearing in the preconditions and post-conditions of the operators which are represented in the plan steps, $\mathcal{S}$.

3. $\mathcal{O}$ is a partial ordering relation on $\mathcal{S}$, representing the ordering constraints over the steps in $\mathcal{S}$.

4. $\mathcal{L}$ is a set of causal links of the form $\langle s, p, s' \rangle$ where $s, s' \in \mathcal{S}$.

5. $\mathcal{E}$ contains step effects, represented as $\langle s, e \rangle$, where $s \in \mathcal{S}$.

6. $\mathcal{C}$ is a set of open conditions of the partial plan, each of which is a tuple $\langle p, s \rangle$ such that $p$ is a precondition of step $s$ and there is no link supporting $p$ at $s$ in $\mathcal{L}$.

Planning consists of starting with a *null plan* (denoted by $\mathcal{P}_\emptyset$), whose candidate set corresponds to all possible ground operator sequences, and successively refining the plan by adding constraints until a solution is reached. Each planning decision represents a choice as to how to resolve an existing flaw in the plan, which is either an open condition (unachieved goal) or a threat to a causal link. To understand how these choices are validated during the replay process it is useful to think of a planning decision as an operator acting on a partly-constructed plan. The possible choices available to DERSNLP+EBL are shown in Figure 6.

Planning decisions have preconditions which are based on the existence of a flaw in the current active plan and have effects which alter the constraints so as to eliminate the flaw. For example, the precondition of an establishment choice is specified in terms of the existence of an unachieved subgoal. The effect is the addition of a causal link that achieves this open condition. The precondition of a resolution decision is a threat in that one step is clobbering an existing causal link. A threat is resolved by adding a step ordering which either promotes or demotes the clobberer relative to the causal link.

Figure 6: Planning decisions are based on the active plan $\langle \mathcal{S}, \mathcal{O}, \mathcal{B}, \mathcal{L}, \mathcal{E}, \mathcal{C} \rangle$ and have effects which alter the constraints so as to produce the new current active plan $\langle \mathcal{S}', \mathcal{O}', \mathcal{B}', \mathcal{L}', \mathcal{E}', \mathcal{C}' \rangle$.

Before a decision is replayed, it is first compared with the current active plan to determine whether its precondition holds in the new context. Invalid decisions, those whose preconditions don't match, are skipped. Establishment decisions are ignored if the goals they achieve are not present as open conditions in the current active plan. Threat resolutions are skipped if the threat is not present. Previous choices which are justified in the current situation are used as guidance to direct the new search process. Replaying a valid decision involves selecting a match for that decision from the children of the current active plan, and making this child the next plan refinement.

DERSNLP+EBL's eager derivation replay strategy replays all of the applicable decisions in the trace in sequence. This replay strategy can be contrasted with that of PRODIGY/ANALOGY (Veloso, 1994) where replay is alternated with from-scratch planning for extra goals not covered by a case. In eager derivation replay each previous decision is eagerly adopted if justified in the current context. Since invalid instructions have been skipped, the skeletal plan which is the end result of replay is comparable to the product of the fitting phase in plan reuse (Kambhampati & Hendler, 1992; Hanks & Weld, 1995). In contrast to plan reuse, derivation replay does not alter the underlying planning strategy. Replay merely provides search control, directing the search as to which node to visit next. This means that DERSNLP+EBL inherits all of the properties of SNLP, including soundness, completeness, and systematicity.

A sample trace of SNLP's decision process is shown in Figure 7. The trace corresponds to a simple problem from the logistics transportation domain of (Veloso, 1994) adapted for SNLP as in Figure 4. This problem contains the goal of getting a single package, OB1, to a designated airport, $l_d$. The derivation trace contains the choices that were made along the path from the root of the search tree to the final plan in the leaf node. Instructions contain a description of both the decision taken and its basis for justification in a new context.

## 2.2 Eager Case Extension and Recovery

The decisions in the trace that are skipped during replay are only those that are known *a priori* to be unjustified. This does not guarantee that the skeletal plan which is left will

Goal : (AT-OB OB1 $l_d$)
Initial : ((IS-A AIRPORT $l_d$) (IS-A AIRPORT $l_i$))
(IS-A AIRPORT $l_p$) (AT-PL PL1 $l_p$)
(AT-OB OB1 $l_i$) ....

| |
|---|
| Name : G1<br>Type : START-NODE |
| Name : G2<br>Type : ESTABLISHMENT<br>Kind : NEW STEP<br>New Step: (UNLOAD-PL OB1 ?P1 $l_d$)<br>New Link: (1 (AT-OB OB1 $l_d$) GOAL)<br>Open Cond: ((AT-OB OB1 $l_d$) GOAL) |
| Name : G3<br>Type : ESTABLISHMENT<br>Kind : NEW STEP<br>New Step: (FLY-PL ?P1 ?A2 $l_d$)<br>New Link: (2 (AT-PL ?P1 $l_d$) 1)<br>Open Cond: ((AT-PL ?P1 $l_d$) 1) |
| Name : G4<br>Type : ESTABLISHMENT<br>Kind : NEW STEP<br>New Step: (FLY-PL ?P1 ?A3 ?A2)<br>New Link: (3 (AT-PL ?P1 ?A2) 2)<br>Open Cond: ((AT-PL ?P1 ?A2) 2) |
| Name : G5<br>Type : ESTABLISHMENT<br>Kind : NEW LINK<br>New Link: (0 (AT-PL ?P1 $l_p$) 3)<br>Open Cond: ((AT-PL ?P1 ?A3) 3) |
| Name : G6<br>Type : ESTABLISHMENT<br>Kind : NEW LINK<br>New Link: (0 (IS-A AIRPORT $l_i$) 3)<br>Open Cond: ((IS-A AIRPORT ?A2) 3) |
| Name : G7<br>Type : ESTABLISHMENT<br>Kind : NEW LINK<br>New Link: (0 (IS-A AIRPORT $l_d$) 2)<br>Open Cond: ((IS-A AIRPORT $l_d$) 2) |
| Name : G8<br>Type : ESTABLISHMENT<br>Kind : NEW STEP<br>New Step: (LOAD-PL OB1 PL1 ?A4)<br>New Link: (4 (INSIDE-PL OB1 PL1) 1)<br>Open Cond: ((INSIDE-PL OB1 PL1) 1) |
| Name : G9<br>Type : ESTABLISHMENT<br>Kind : NEW LINK<br>New Link: (3 (AT-PL PL1 $l_i$) 4)<br>Open Cond: ((AT-PL PL1 ?A4) 4) |
| Name : G10<br>Type : RESOLUTION<br>Kind : PROMOTION<br>Unsafe-link : ((3 (AT-PL PL1 $l_i$) 4)<br>Effect : 2 ¬(AT-PL PL1 $l_i$)) |
| Name : G11<br>Type : ESTABLISHMENT<br>Kind : NEW LINK<br>New Link: (0 (AT-OB OB1 $l_i$) 4)<br>Open Cond: ((AT-OB OB1 $l_i$) 4) |

Key to Abbreviations:
PL = PLANE
OB = OBJECT

Final Plan: (FLY-PL PL1 $l_p$ $l_i$) Created 3
(LOAD-PL OB1 PL1 $l_i$) Created 4
(FLY-PL PL1 $l_i$ $l_d$) Created 2
(UNLOAD-PL OB1 PL1 $l_d$) Created 1
Ordering of Steps: ((4 ¿ 2) (3 ¿ 4) (4 ¿ 1) (3 ¿ 2) (2 ¿ 1))

Figure 7: An Example solution trace for DERSNLP+EBL

$$\theta_2 D^m S^1 :$$
$$(A_i^\alpha \ \underline{precond} : \ \{I_i, P_\alpha\} \ \underline{add} : \ \{g_i\} \ \underline{delete} : \ \{I_j | j < i\})$$
$$(A_i^\beta \ \underline{precond} : \ \{I_i P_\beta\} \ \underline{add} : \ \{g_i\} \ \underline{delete} : \ \{I_j | j < i\})$$
$$(A_\alpha \ \underline{precond} : \ \{\} \ \underline{add} : \ \{g_\alpha\} \ \underline{delete} : \ \{P_\beta\} \cup \{g_i | \forall i\})$$

Figure 8: The specification of Barrett and Weld's Transformed $D^m S^1$ Domain

ultimately be refined into a solution for the current problem. Without actually completing the search there is no way of predicting whether the constraints that are left in the skeletal plan are consistent with a complete solution. Whenever the skeletal plan is not complete (whenever there are extra goals or unsatisfied initial state conditions) the planner must undergo further planning effort to extend the plan and there is a possibility that this effort may fail, necessitating a recovery phase.

For DERSNLP+EBL, the skeletal plan is extended first, prior to recovery. This plan is backtracked over only after the search process fails to refine it into a full solution to the new problem. This strategy requires a depth limit to be placed on the search tree[2]. Otherwise skeletal plan extension may continue indefinitely, and the planning algorithm becomes incomplete. An eager extension strategy is not, however, linked to a particular search method. For example, it may be used with best-first, depth-first or iterative deepening search. These different search methods are used in the exploration of the subtree under the skeletal plan, prior to backtracking over this plan. Once the skeletal plan is found to fail, the recovery phase that is initiated merely involves exploring the siblings of the replayed path. Like extension, recovery is not linked to a particular search strategy.

## 2.3 Analyzing the Failure of Case Extension

In order for the skeletal plan to be successfully extended to achieve any conditions left open, the sequence of decisions that were adopted through the guidance of a previous trace must be concatenated with further choices to arrive at a solution. For this to occur, the replayed path must be *decision-sequencable* with respect to the new problem, which is defined as follows:

**Definition 1 (Decision-Sequencable Search Path)** *A search path which contains a sequence of decisions $D$ is decision-sequencable with respect to a new problem, $\langle I', G', \mathcal{A} \rangle$, if and only if there exist two decision sequences $E$ and $E'$ such that $E \bullet D \bullet E'$ (where "$\bullet$" is the decision sequencing operator) will produce a plan which is correct for $\langle I', G', \mathcal{A} \rangle$.*

One of the primary reasons a replayed path may not be decision sequencable is the goal interactions that occur between the input goals of the new problem. In particular, the extra goals not achieved by a case may interact with those that are covered, making the retrieved case inapplicable. It has long been recognized that the relative difficulty of problem-solving is linked to the level of interaction between the various input goals of the problem (Korf, 1987; Joslin & Roach, 1990; Barrett & Weld, 1994; Veloso & Blythe, 1994; Kambhampati,

---

2. In practice, this limit is actually a bound placed on the number of steps contained in the plan.

Ihrig, & Srivastava, 1996a). Goal interaction has been formalized by Korf (1987) in terms of the problem search space. Barrett and Weld (1994) extend Korf's analysis into plan space. For the plan-space planner, the order in which goals are achieved is not as crucial. Goals that are *laboriously serializable* for a state-space planner (in that there exist few goal orderings for which the goals may be solved in sequence) may be *trivially serializable* for a plan-space planner (meaning that the goals can be solved in any order).

However, goals are not always trivially serializable for the plan-space planner (Veloso & Blythe, 1994). For example, consider the $\theta_2 D^m S^1$ domain (Barrett & Weld, 1994) shown in Figure 8. Notice that if $g_\alpha$ is one of a set of problem goals, and it is not true initially, then any other goal, $g_i$, that is present in the set must be achieved by the operator $A_i^\alpha$, and not by $A_i^\beta$. This means that any time a case is replayed that previously solved a goal, $g_i$, through an action $A_i^\beta$, and $g_\alpha$ is an extra goal not covered by the case, replay will fail.

In CBP, however, we are not so much concerned with the general properties of the domain, as we are with properties of the particular search paths which are stored in the case library. It is not required that the input goals of every problem be trivially serializable for CBP to be beneficial to planning performance. If it were, there would be very few domains in which CBP was effective. Trivial serializability is not a requirement since it is not necessary that every plan for every subset of the input goals be consistent with a solution to the full problem. It is only the particular plans that are retrieved from the library that we are concerned with.

Even if the goals of the problem are not trivially serializable, replay may be decision sequencable, depending on which cases are actually retrieved from the library. In the $\theta_2 D^m S^1$ domain, if the single-goal cases that are retrieved solve $g_i$ through the action $A_i^\beta$, then these will not be decision-sequenceable with any new multi-goal problem which contains the goal $g_\alpha$. However if the stored cases are solved through $A_i^\alpha$, then replay of these cases will be sequencable. In fact, the aim of the DERSNLP+EBL's learning component is to achieve an indexing within the case library such that all of the new problems encountered by the planner may be solved through sequenced replay of the cases retrieved from that library. The next section describes how DERSNLP+EBL is able to work towards this objective through its learning component which learns from replay failures.

## 2.4 Constructing Reasons for Retrieval Failure

DERSNLP+EBL constructs explanations for retrieval failures through the use of explanation-based learning techniques which allow the planner to explain the failures of individual plans in the planner's search space. A leaf node plan represents an analytical failure when it contains a set of inconsistent constraints which prevent the plan from being further refined into a solution. An analytical failure is explained in terms of these constraints (Kambhampati, Katukam, & Qu, 1996b). Leaf node explanations identify a minimal set of constraints in the plan which are together inconsistent. DERSNLP+EBL forms explanations for each of the analytical failures that occur in the subtree directly under the skeletal plan. These are regressed up the failing search paths and are collected at the root of the tree to form a reason for the retrieval failure (See Figure 9a). The regressed explanation is in terms of the new problem specification. It contains a subset of interacting goals, as well as initial state conditions relevant to those goals.

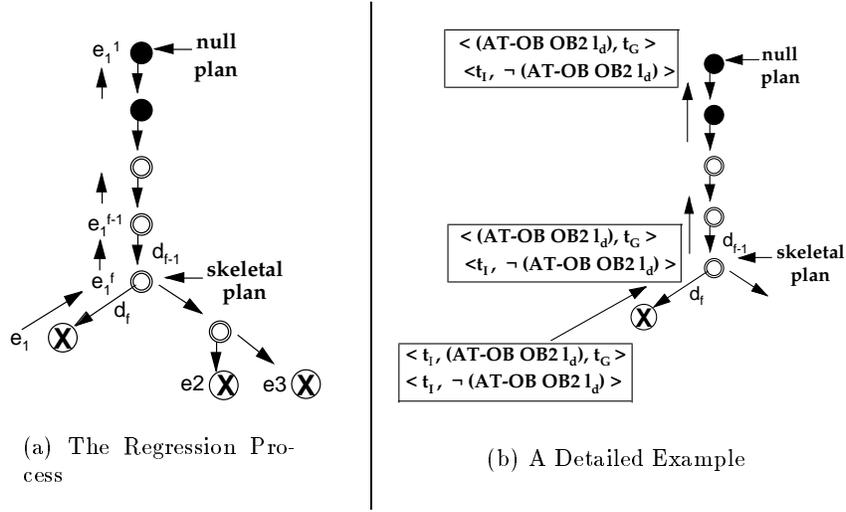

(a) The Regression Process

(b) A Detailed Example

Figure 9: The path failure explanation at the root of the tree is computed as $e_1^1 = d_1^{-1}(d_2^{-1} \cdot \cdots (d_f^{-1}(e_1)) \cdots)$.

Since a plan failure is explained by a subset of its constraints, failure explanations are represented in the same manner as the plan itself. Recall that DERSNLP+EBL represents its plans as a 6-tuple, $\langle \mathcal{S}, \mathcal{O}, \mathcal{B}, \mathcal{L}, \mathcal{E}, \mathcal{C} \rangle$ (See Section 2). The explanation for the failure occurring at a leaf node contains only the constraints which contribute to an inconsistency. These inconsistencies appear when new constraints are added which conflict with existing constraints. As discussed in Section 2, DERSNLP+EBL makes two types of decisions, *establishment* and *resolution*. Each type of decision may result in a plan failure. An *establishment* decision represents a choice as to a method of achieving an open condition, either through a new/existing step, or by adding a causal link from the initial state. When an attempt is made to achieve a condition by linking to an initial state effect, and this condition is not satisfied in the initial state, the plan then contains a contradiction. An explanation for the failure is constructed which identifies the two conflicting constraints:

$$\langle \emptyset, \emptyset, \emptyset, \{\langle t_I, p, s \rangle\}, \{\langle t_I, \neg p \rangle\}, \emptyset \rangle$$

The precondition of a resolution decision is a threat to a causal link. DERSNLP+EBL uses two methods of resolving a threat, promotion and demotion, each of which adds a step ordering to the plan. When either decision adds an ordering which conflicts with an existing ordering, an explanation of the failure identifies the conflict:

$$\langle \emptyset, \{s \prec s', s' \prec s\}, \emptyset, \emptyset, \emptyset, \emptyset \rangle$$

Each of the conflicting constraints in the failure explanation is regressed through the final decision, and the results are sorted according to type to form the new regressed explanation. This process is illustrated graphically in Figure 9b. In this example, a new link from the initial state results in a failure. The explanation, $e_1$ is:

$$\langle \emptyset, \emptyset, \emptyset, \{\langle t_I, (AT-OB\ OB2\ l_d), t_G \rangle\}, \{\langle t_I, \neg(AT-OB\ OB2\ l_d) \rangle\}, \emptyset \rangle$$

When $e_1$ is regressed through the final decision, $d_f$, to obtain a new explanation, the initial state effect regresses to itself. However, since the link in the explanation was added by the decision, $d_f$, this link regresses to the open condition which was a precondition of adding the link. The new explanation, $e_1^f$, is therefore

$$\langle \emptyset, \emptyset, \emptyset, \emptyset, \{\langle t_I, \neg(AT\text{--}OB\ OB2\ l_d)\rangle\}, \{\langle(AT\text{--}OB\ OB2\ l_d), t_G\rangle\}\rangle$$

The regression process continues up the failing path until it reaches the root of the search tree. When all of the paths in the subtree underneath the skeletal plan have failed, the failure reason at the root of the tree provides the reason for the failure of the retrieved cases. It represents a combined explanation for all of the path failures. The case failure reason contains only the aspects of the new problem which were responsible for the failure. It may contain only a subset of the problem goals. Also, any of the initial state effects that are present in a leaf node explanation, are also present in the reason for case failure[3].

## 2.5 An Empirical Evaluation of the Utility of Case Failure Analysis

A preliminary study was conducted with the aim of demonstrating the advantage of storing and retrieving cases on the basis of experienced retrieval failure. Domains were chosen in which randomly generated problems contained negatively interacting goals, and planning performance was tested when DERSNLP+EBL was solving multi-goal problems from scratch and through replay of single cases covering a smaller subset of goals. Replay performance was tested both *with* and *without* case failure information.

### 2.5.1 Domains

Experiments were run on problems drawn from two domains. The first was the artificial domain, $\theta_2 D^m S^1$, originally described in (Barrett & Weld, 1994) and shown in Figure 8. Testing was done on problems which were randomly generated from this domain with the restriction that they always contain the goal $g_\alpha$. The Logistics Transportation domain of (Veloso, 1994) was adopted for the second set of experiments. Eight packages and one airplane were randomly distributed over four cities. Problem goals represented the task of getting one or more packages to a single destination airport[4]. The FLY operator was augmented with a delete condition which prevented planes from visiting the same airport more than once. This meant that replay failed if there was an extra package to be transported which was off the previous route taken by the plane.

### 2.5.2 Retrieval Strategy

Cases were initially retrieved on the basis of a *static* similarity metric which takes into account the goals that are covered by the case as well as all of their relevant initial state conditions (Kambhampati, 1994; Veloso, 1994). Prior studies show it to be a reasonably effective metric. In *learning* mode, cases were also retrieved on the same basis. However, in this mode, the failure reasons attached to the case were used to censor its retrieval. Each time that a case was retrieved in learning mode, these failure conditions were also tested. If each failure reason was not satisfied in the new problem specification, the retrieval mechanism returned the case for replay. If, on the other hand, a failure reason was found to be true in the new problem context, then the case that repaired the failure was retrieved. Following retrieval, the problem was solved both by replay of the retrieved case as well as by planning from scratch.

---

3. DERSNLP+EBL's EBL component explains only analytical failures. Depth limit failures are ignored. This means that the failure explanations that are formed are not sound in the case of a depth limit failure, and that the retriever may reject a case when it is applicable. Rejecting an applicable case may lead to the storage of duplicate cases and a larger library size. However, our empirical work has not shown this to be of practical importance for reasons outlined in Section 3.2.2.

4. For a more comprehensive evaluation over an unbiased problem set see Section 4.

### 2.5.3 EXPERIMENTAL SETUP

Each experiment consisted of three phases, each phase corresponding to an increase in problem size. Goals were randomly selected for each problem, and, in the case of the logistics domain, the initial state was also randomly varied between problems. In an initial training session that took place at the start of each phase $n$, 30 $n$-goal problems were solved from scratch, and each derivation trace was stored in the library. Following training, the testing session consisted of generating problems in the same manner but with an additional goal. Each time that a new $(n+1)$ goal problem was tried, an attempt was made to retrieve a similar $n$-goal problem from the library. If during the testing session, a case that was similar to the new problem was found which had previously failed, then the problem was solved in learning, static and from-scratch modes, and it became part of the 30-problem set. With this method, we were able to evaluate the improvements provided by failure-based retrieval when retrieval on the static metric alone was ineffective, and when failure conditions were available.

### 2.5.4 EXPERIMENTAL RESULTS

The results of the experiments are shown in Tables 1 and 2. Each table entry represents cumulative results obtained from the sequence of 30 problems corresponding to one phase of the experiment. The first row of Table 1 shows the percentage of problems correctly solved within the time limit (550 seconds). The average solution length is shown in parentheses for the logistics domain (solution length was omitted in $\theta_2 D^m S^1$ since all of the problems generated within a phase have the same solution length). The second and third rows of Table 1 contain respectively the total number of search nodes visited for all of the 30 test problems, and the total CPU time (including case retrieval time).

These results are also summarized in Figure 10. DERSNLP+EBL in learning mode was able to solve as many of the multi-goal problems as in the other two modes and did so in substantially less time. Case retrieval based on case failure resulted in performance improvements which increased with problem size. Comparable improvements were not found when retrieval was based on the static similarity metric alone. This should not be surprising since cases were retrieved that had experienced at least one earlier failure. This meant that testing was done on cases that had some likelihood of failing if retrieval was based on the static metric.

Table 2 records three different measures which reflect the effectiveness of replay. The first is the percentage of *sequenced* replay. Recall that replay of a trace is considered here to be sequenced if the skeletal plan is further refined to reach a solution to the new problem. The results point to the greater efficiency of replay in learning mode. In the $\theta_2 D^m S^1$ domain, replay was entirely sequenced in this mode. In the transportation domain, retrieval based on failure did not always result in sequenced replay, but did so more often than in static mode.

| Phase | $\theta_2 D^m S^1$ | | | Logistics | | |
|---|---|---|---|---|---|---|
| | **Learning** | **Static** | **Scratch** | **Learning** | **Static** | **Scratch** |
| **(1) Two Goal** | | | | | | |
| %Solved | 100% | 100% | 100% | 100% (6.0) | 100% (6.0) | 100% (6.0) |
| nodes | 90 | 240 | 300 | 1773 | 1773 | 2735 |
| time(sec) | 1 | 4 | 2 | 30 | 34 | 56 |
| **(2) Three Goal** | | | | | | |
| % Solved | 100% | 100% | 100% | 100% (8.2) | 100% (8.2) | 100% (8.2) |
| nodes | 120 | 810 | 990 | 6924 | 13842 | 20677 |
| time(sec) | 2 | 15 | 8 | 146 | 290 | 402 |
| **(3) Four Goal** | | | | | | |
| % Solved | 100% | 100% | 100% | 100% (10.3) | 100% (10.3) | 100% (10.3) |
| nodes | 150 | 2340 | 2533 | 290 | 38456 | 127237 |
| time(sec) | 3 | 41 | 21 | 32 | 916 | 2967 |

Table 1: Performance statistics in $\theta_2 D^m S^1$ and Logistics Transportation Domain (Average solution length is shown in parentheses next to %Solved for the logistics domain only)

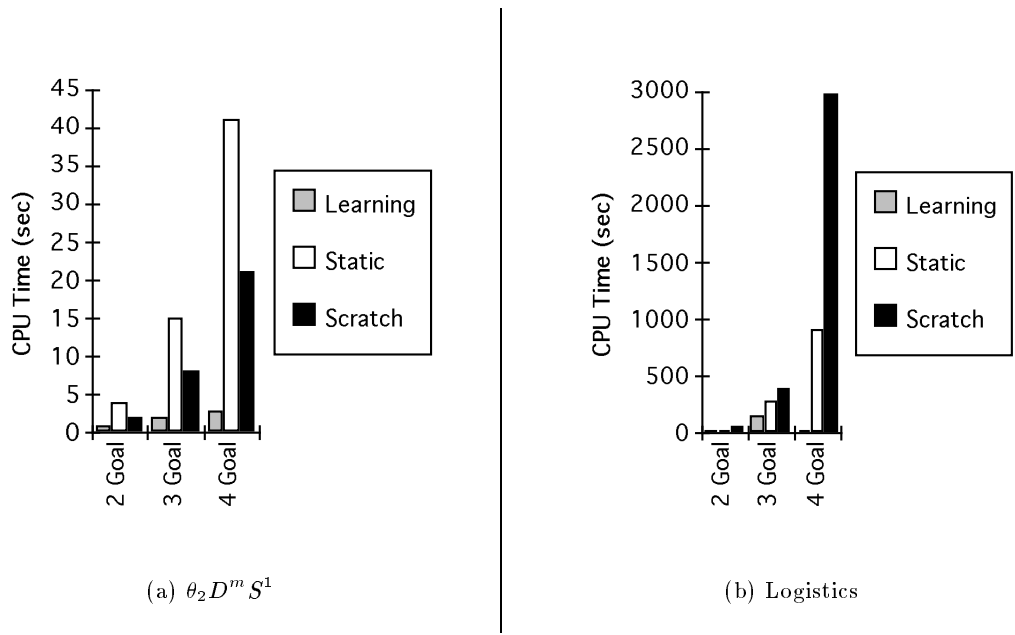

(a) $\theta_2 D^m S^1$

(b) Logistics

Figure 10: Replay performance in the $\theta_2 D^m S^1$ and Logistics Transportation domain.

| Phase | $\theta_2 D^m S^1$ | | Logistics | |
|---|---|---|---|---|
| | **Learning** | **Static** | **Learning** | **Static** |
| **Two Goal** | | | | |
| % Seq | 100% | 0% | 53% | 53% |
| % Der | 60% | 0% | 48% | 48% |
| % Rep | 100% | 0% | 85% | 85% |
| **Three Goal** | | | | |
| % Seq | 100% | 0% | 80% | 47% |
| % Der | 70% | 0% | 63% | 50% |
| % Rep | 100% | 0% | 89% | 72% |
| **Four Goal** | | | | |
| % Seq | 100% | 0% | 100% | 70% |
| % Der | 94% | 0% | 79% | 62% |
| % Rep | 100% | 0% | 100% | 81% |

Table 2: Measures of effectiveness of replay.

The greater effectiveness of replay in learning mode is also indicated by the two other measures contained in the subsequent two rows of Table 2. These are respectively, the percentage of plan refinements on the final derivation path that were formed through guidance from replay (% Der), and the percentage of the total number of plans created through replay that remain in the final derivation path (% Rep). The case-based planner in learning mode showed as much or greater improvements according to these measures, demonstrating the relative effectiveness of guiding retrieval through a learning component based on replay failures. These results indicate that DERSNLP+EBL's integration of CBP and EBL is a promising approach when extra interacting goals hinder the success of replay.

In Section 4 we report on a more thorough evaluation of DERSNLP+EBL's learning component. This was conducted with the purpose of investigating if learning from case failure is of benefit for a planner solving random problems in a complex domain. For this evaluation we implemented the full case-based planning system along with novel case storage and adaptation strategies. In the next section, we describe the storage strategy that was developed for this evaluation.

## 3. Improving Case Storage and Adaptation

The aim of case-based planning is to efficiently solve large problems in complex domains. A complex domain means a great variety in the problems encountered. When problem size (measured in terms of the number of goals, $n$) is large, it is unlikely that the same $n$-goal problem will have been seen before. It is therefore an advantage to be able to store cases covering smaller subsets of goals, and to retrieve and adapt multiple cases in solving a single large problem.

Before implementing this strategy, decisions had to be made as to which goal combinations to store. In previous work within state-space planning Veloso (1994) has developed an approach to reducing the size of the library by first transforming a totally ordered plan into a partially ordered graph, separating out connected components of the graph, and storing these subplans individually. Goals which interact in that their respective plans must be interleaved in order to form a complete solution are stored together as a single case. When replay is based on a plan-space planner such as SNLP, a component subplan may be further subdivided, since the planner has the ability to first piece plans together, and later add step orderings to interleave these subplans (Kambhampati & Chen, 1993; Ihrig & Kambhampati, 1994a). Replay of these smaller cases will be sequenced as long as their individual subplans may be interleaved by the addition of step orderings to form a full solution. The plan-space planner therefore has a greater capability of reducing the size of the problems stored in the library, and, as a consequence, the number of cases stored.

DERSNLP+EBL's storage strategy makes use of the plan-space planners' ability to piece small plans together, then add step orderings to interleave these plans. As in earlier approaches, such as PRIAR (Kambhampati & Hendler, 1992), PRODIGY/ANALOGY (Veloso, 1994) and CAPLAN (Munoz-Avila & Weberskirch, 1996), the cases that are stored cover smaller subsets of the original set of input goals achieved in the successful problem-solving episode. DERSNLP+EBL differs from these earlier approaches in that the division into goal subsets is not based on the structure of the final plan alone, but on the sequence of events making up the problem-solving episode. A new repairing case is stored if the cases which were retrieved from the library in solving the new problem fail to be extended into a new solution. The storer constructs a new case based on the failure explanation which was obtained through the extension phase as well as the new successful plan derivation obtained during recovery.

The failure explanation identifies the set of negatively interacting goals responsible for the failure. These goals form a subset of the input goals which are achieved in the new solution. Before the repairing case is stored, the new plan derivation is stripped of any decisions that are irrelevant to the achievement of these interacting goals. The new case then covers only the negatively interacting goals.

Note that we define negative interaction based on the failure of the skeletal plan. An interaction occurs when a set of input goals cannot be solved by refining the skeletal plan, causing the planner to have to backtrack over this plan. Moreover, we cannot determine whether two goals are negatively interacting merely by analyzing the final solution. It does not include information about the planning failures which were encountered in generating the solution. In particular, the final solution does not tell us whether an additional goal was achieved by extending the replayed path, or by backtracking over that path. Approaches to case storage which determine goal interaction from the final plan alone (Veloso, 1994; Munoz-Avila & Weberskirch, 1996) therefore ignore the retrieval failures that have been encountered during the planning episode.

Retrieval failures provide important guidance as to how the library may be improved to avoid similar failures. For DERSNLP+EBL, they are used to dynamically improve the storage in the library through the addition of new goal combinations. Multi-goal problems are stored when retrieved cases corresponding to single-goal subproblems fail to be merged and

---

Case Failure Explanation:

$\mathcal{C} = \{\langle g_\alpha, t_G \rangle, \langle g_8, t_G \rangle\}$
$\mathcal{E} = \{\langle t_I, i_8 \rangle, \langle t_I, P_\beta \rangle\}$

---

Figure 11: An Example of a case failure reason

extended into a new solution. Repairing cases are constructed which achieve the negatively interacting goals which are responsible and which are identified in the failure explanation.

## 3.1 An Example of a Negative Interaction

Figure 11 provides an example of an explanation for failure encountered when solving a problem from Barrett and Weld's $\theta_2 D^m S^1$ domain shown in figure 8. The problem contains three goals, $g_\alpha$, $g_6$ and $g_8$, and has been attempted through replay of a case which solves two of these goals, $g_\alpha$ and $g_6$, and a second case, which achieves only $g_8$. In the latter, the goal was achieved through the action $A_8^\beta$ which represents an incorrect operator choice when the input goals of the problem include the goal $g_\alpha$.

The failure explanation shown in Figure 11 identifies a subset of interacting goals, made up of $g_8$ and $g_\alpha$. Note that this interaction is not evident in the final plan shown in Figure 12. In this plan, the three input goals of the problem are achieved through the same connected component. If we base storage solely on the plan graph represented by the successful plan, then all three input goals will be stored in a single case. Moreover, each new problem representing a novel combination of goals will be stored in the library, causing the library size to increase exponentially with problem size. For example, suppose the domain includes the goals, $\{g_i | 1 < i < n\}$ and $g_\alpha$. Then the number of problems of size three will be the number of 3-goal subsets of these $n + 1$ goals. DERSNLP+EBL's strategy of storing cases based on explanations of retrieval failure will result in a maximum of $2n + 1$ cases stored. Each goal $\{g_i | 1 < i < n\}$ appears in only two cases, one representing the single-goal problem and one representing a two goal problem which also achieves $g_\alpha$.

Storing only negatively interacting goals as multi-goal problems may therefore result in a substantial reduction in the size of the case library. It also represents a tradeoff, as the replayed cases must be extended by from-scratch planning to solve conflicts between the individual plans recommended by separate cases. Moreover, in more complex domains, there may be goals which interact *positively* in that they may be solved through common steps (Ihrig & Kambhampati, 1996; Munoz-Avila & Weberskirch, 1997). If these goals are stored as separate cases, then replay may result in unnecessary redundancy in the plan. In DERSNLP+EBL, these positive interactions are handled through the replay process itself, which merges the subplans provided by multiple cases. In Section 3.3 we describe how this merging is accomplished. The next section provides more detail as to the case storage strategy which has been implemented for our empirical study.

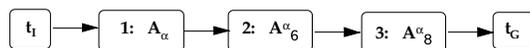

Figure 12: A Solution to the example problem.

## 3.2 Building the Case Library

The following deliberative strategy was adopted for building the case library. When a new problem contains $n$ goals, the first goal is attempted, and, if solved, the case covering this goal alone is stored in the library. Problem-solving continues by increasing the problem size by one goal at a time. For example, if the problem just attempted contained the goal set, $G = \langle g_1, g_2, ..., g_i \rangle$ and was solved through a decision sequence $D_i$ then a second decision sequence, $D_{i+1}$, is stored whenever $D_i$ cannot be replayed and extended to achieve the next goal $g_{i+1}$. Whenever the replayed derivation path fails, and the recovery phase is successful in producing a new solution, the explanation for the case retrieval failure is used to identify a subset of negatively interacting input goals, $N = \langle g_j...g_{j+m} \rangle$, that are responsible for the failure. If the replayed path fails to be extended, and is backtracked over to reach a solution to the new problem, then the new successful derivation is passed to the *storer* along with the failure explanation. The explanation is used to delete from the derivation any decisions which are not relevant to the set of negatively interacting goals, $N$. This reduced derivation is then stored in the library as the repairing case. Alternatively, whenever the next goal in the set is solved through simple extension of the previous decision sequence, no case is stored which includes that goal.

This storage strategy entails two important properties. (1) Each new case corresponds to either a new single-goal problem or to a multi-goal problem containing *negatively interacting* goals. (2) All of the plan derivations arising from a single problem-solving episode are different in that no decision sequence stored in the library is a prefix of another stored case. This is because no case is added to the library when a new problem is solved by extending a retrieved case. New cases are stored only when some of the previous decisions need to be backtracked over in the search for a new solution.

DERSNLP+EBL's strategy of restricting multi-goal cases to those with goals which are negatively interacting serves to ameliorate the mis-retrieval problem. The more experience that the planner has in problem-solving, the more of these interactions are discovered, and the less likely it is that the planner has to backtrack over its replayed paths. The aim is to eventually have in the library a minimal number of cases such that all of the problems encountered may be achieved by successfully merging multiple instances of stored cases. The approach is therefore to retain cases based on their *competence* as well as their performance (Smyth & Keane, 1995).

### 3.2.1 An Example of DERSNLP+EBL's Storage Strategy

As an example of how a multi-goal problem is stored, consider the problem contained in Figure 13 where three packages, OB1, OB2 and OB3, are to be transported to the same destination location, $l_d$. Initially the goal set contains the goal of transporting OB1 alone, represented as (AT-OB OB1 $l_d$), and the successful derivation is stored as Case $A$. The second goal is then added to the set. Since the problem just attempted achieves the first goal through a decision sequence which has to be backtracked over in order to solve the additional goal, a second derivation, Case $B$, is stored. This new derivation then solves the mutually interacting goals, (AT-OB OB1 $l_d$) and (AT-OB OB2 $l_d$). Problem-solving then continues with the addition of the third goal. This goal is solved through simple extension of the previous decision sequence. No case is stored which includes this goal. This means that we have two cases stored in the library: Case $A$ corresponding to a single-goal problem and Case $B$ corresponding to a multi-goal problem containing two negatively interacting goals. Multi-goal problems are stored only when the problem goals are mutually interacting, that is, only when their individual derivations cannot be sequenced and extended to solve the full problem.

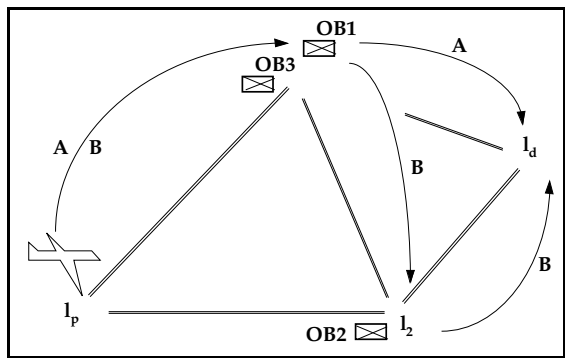

Figure 13: A logistics transportation example illustrating multi-case storage. The figure shows two plans produced by two stored derivations. Case $A$ achieves the goal of having a single packages, OB1, transported to the destination airport, $l_d$. Case $B$ achieves the goal of having OB1 and OB2 located at the same airport.

With DERSNLP+EBL's storage strategy, the size of the library is limited by the amount of interaction in the domain. For example, if there is no negative interaction, then only single goal cases will be stored. In the logistics transportation domain, there is a potential for all problem goals to interact negatively. However, since there are also a significant percentage of non-interacting goals, this strategy reduces the size of the library in comparison to one in which all of the multi-goal problems which are successfully solved are stored. This storage strategy also represents a tradeoff since effort must be expended in merging the retrieved cases into a full solution (See Section 3.3).

### 3.2.2 INDEXING ON THE BASIS OF REPLAY FAILURE

Multi-goal cases are stored in the library so as to censor the retrieval of their corresponding single-goal subproblems. This library organization differs from earlier work which stores all cases in a common fashion on a single level, first indexing each case by all of the goals, then by all of the success conditions relevant to these goals (Veloso, 1994; Munoz-Avilla & Weberskirch, 1996). In contrast, DERSNLP+EBL indexes its cases through a discrimination net similar to the one depicted in Figure 14. This figure shows one fragment of the case library which includes all of the cases which solve a single input goal. Individual planning episodes which achieve this goal are represented one level lower in the net. Each is labeled by its relevant initial state conditions, otherwise known as the footprinted initial state (Veloso, 1994). Together, the goal and initial state conditions make up the static success conditions on which cases are first retrieved. When one of these cases is retrieved for replay and replay fails, the derivation corresponding to the extra interacting goals is added to the library and indexed directly under the failing case. On future retrievals of the case, the failure conditions are checked to see whether the extra goals responsible for the failure are present under the same conditions. If so, the retrieval process returns the repairing case which achieves these conflicting goals. The case failure reason is thus used to direct retrieval away from the case which will repeat a known failure, and towards the case that avoids it.

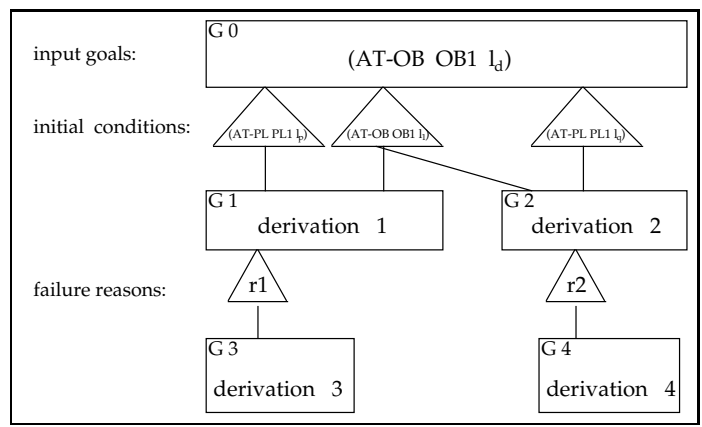

Figure 14: A Library fragment indexing stored cases which solve a single input goal, (AT-OB OB1 $l_d$).

One might question this hierarchical organization in instances where failures are due to interacting goals alone. Why not just store all cases on a single level first indexing each case by all of its goals, then by the conditions relevant to all of these goals? The answer lies in the need to censor cases when failure conditions are satisfied. This type of error will be found when retrieving multiple cases. As an example, consider that our new problem contains three goals, $g_1$, $g_2$ and $g_3$. Suppose further that the goal $g_2$ negatively interacts with both $g_1$ and $g_3$. If a case is retrieved from the library which achieves both $g_1$ and $g_2$, then one goal, $g_3$, is left open. However, if a case is then retrieved which solves $g_3$ alone, it will fail because of the presence of $g_2$. This type of retrieval error is handled by *prioritizing* cases. A repairing case is stored as a subclass of the case that failed. Failing cases are annotated with the failure reason which directs the retriever to the case that avoids the failure.

Prioritizing cases on the basis of negatively interacting goals alone is not sufficient to capture all of the retrieval failures that may be encountered. If cases are retrieved on the basis of a partial match of the relevant initial state conditions, then retrieval errors may occur because of unmatched conditions (Veloso, 1994). For example, just as a failure might occur in our logistics transportation example if there is an extra package off the plane's route, a similar failure will occur if a package is moved off the plane's route. The strategy that is adopted to deal with both types of failure information is to annotate the case with the failure reason (whether it is an extra goal or an unmatched initial state condition) and use the failure reasons to prioritize cases. The EBL techniques that we have employed in the construction of failure explanations may be used for both types of failures.

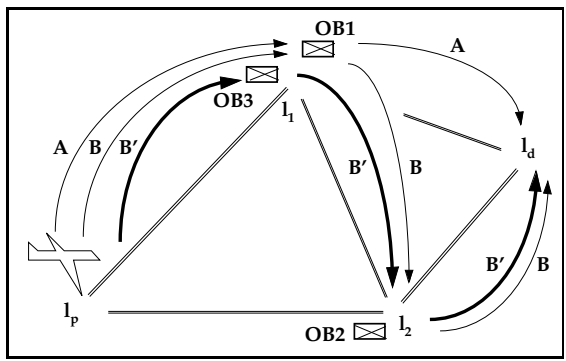

Figure 15: A logistics transportation example illustrating multi-case retrieval.

DERSNLP+EBL's method of storing multi-goal cases only when goals are negatively interacting limits the size of the case library. Other aspects of DERSNLP+EBL's storage strategy also serve to lower library size. The planner always uses its current library in solving new problems. New derivations are stored only when there is no applicable case, or when the retrieved cases fail. This strategy avoids the storage of duplicate cases, but may not be entirely effective since the soundness of failure explanations is not guaranteed. If failure explanations are not sound, pointers to repairing cases may eventually lead to a duplicate case, causing the library to continue to grow indefinitely. However, this is easily checked by putting a depth limit on the number of repairing cases in the discrimination net. Also, failures which are due to interacting goals will not result in unchecked growth of the library since the number of interacting goals is limited by the maximum problem size.

### 3.2.3 A Detailed Example of Case Retrieval

An example of case retrieval is illustrated in Figure 15. The figure contains three subplans corresponding to two separate cases stored in the library. Case $A$ achieves the goal of having a single package, OB1, located at destination $l_d$. Case $B$ achieves the goal of having both OB1 and OB2 located at $l_d$.

Assume that a new problem which is to be attempted through replay contains three goals, (AT-OB OB1 $l_d$), (AT-OB OB2 $l_d$), and (AT-OB OB3 $l_d$). The second goal negatively interacts with both of the other goals. The retriever will first attempt to find a case that solves the first goal alone. Case $A$ solves this goal. However, this case is annotated with a failure reason which is satisfied in the new problem situation, and $A$ is therefore censored in favor of the repairing case, Case $B$. Once the retriever returns Case $B$, it will then have one open goal not covered, that is, (AT-OB OB3 $l_d$). It will seek out a case which solves this goal alone, and will again find Case $A$. However, $A$'s failure reason is again satisfied in the new problem state and will be rejected in favor of a second copy of $B$ (which we now call Case $B'$), which solves the problem of transporting both OB3 and OB2. There will then be two instances of Case $B$ that will be retrieved to solve the three goal problem, Case $B$ and Case $B'$. Together they cover the new problem goals. DERSNLP+EBL replays both copies of $B$ in sequence to obtain a solution to the full problem, thereby merging their respective subplans. Notice, however, that the union of these plans will contain redundant steps. For example, both plans have the plane fly to location $l_1$. Section 3.3 describes how DERSNLP+EBL deals with these positive goal interactions.

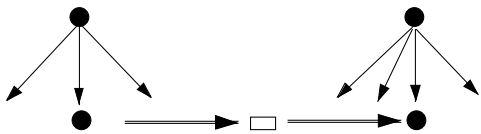

Figure 16: New linking opportunities indicated by an increase in the number of siblings of the step addition decision.

## 3.3 Multi-case Merging

We say that two plans are mergeable with respect to a problem, $\langle I', G', \mathcal{A} \rangle$, if there exists a solution to the problem which contains all of their combined constraints.

**Definition 2 (Mergeability)** *A plan $P_1$ for achieving goal $g_1$ is mergeable with a plan $P_2$ for the goal $g_2$ with respect to a problem, $\langle I', G', \mathcal{A} \rangle$ , if there is a plan $P'$ which is correct for $\langle I', G', \mathcal{A} \rangle$ and $\langle\!\langle P' \rangle\!\rangle \subseteq \langle\!\langle P_1 \rangle\!\rangle \cap \langle\!\langle P_2 \rangle\!\rangle$. (Thus syntactically, $P'$ contains all the constraints of $P_1$ and $P_2$).*

Multi-case replay accomplishes plan merging, but may result in lower quality plans if care is not taken to avoid redundant step additions (Ihrig & Kambhampati, 1996; Munoz-Avilla & Weberskirch, 1997). These occur when goals covered by separate cases *positively interact* in that they may be solved through common steps. Replaying each case in sequence then results in unneeded steps in the plan[5].

In multi-case replay, if an open condition is the only justification for adding a new step, some steps may be added which already exist in the plan due to the earlier replay of another case. When the first retrieved derivation is replayed, none of its replayed step additions will result in redundancy. However, when subsequent goals are solved through replay of additional cases, some step additions may be unnecessary in that there are opportunities for linking the open conditions they achieve to earlier established steps. The planner has no way of determining *a priori* that these steps may be represented by a single step in the plan[6].

DERSNLP+EBL's replay framework handles redundant step additions by skipping over step addition establishments whenever the open condition may be achieved by a new link. It thus strengthens or *increases the justification* for replaying step addition decisions in that the open condition is no longer the only basis for validating the decision. The justification for replay is strengthened to add the condition that no new linking opportunities are present. These may be detected as an increase in the number of siblings of the prescribed step addition choice (See Figure 16). The siblings of the stored step addition decision are recorded as annotations on the derivation trace. When new links are available which are not contained within these siblings, the step addition decision is skipped. After replay, the alternative new links are explored through the normal course of plan refinement. This means that the same step may eventually be added if the new links fail.

---

5. An analogous decrease in plan quality occurs in state-space plan reuse, when sequencing macro-operators results state loops (Minton, 1990a).

6. Consider, for example, a domain in which the plane may transport two packages in one trip, or not, depending on its capacity.

| | DERSNLP+EBL | | DERSNLP+EBL-IJ | |
|---|---|---|---|---|
| | replay | scratch | replay | scratch |
| %Solved | 87%(6) | 67%(5) | 87%(7) | 67% (5) |
| time(sec) | 2465 | 5796 | 2198 | 5810 |

Table 3: Percentage problems solved, total CPU time in seconds on all 30 problems for problems in the Logistics Transportation Domain. Average solution length is shown in parentheses next to %Solved.

Increasing the justification for the step addition decisions improves the quality of plans in terms of the number of steps they contain. For example, Case $B$ and $B'$ would normally produce subplans which are shown in Figure 15. When these cases are replayed in sequence in solving a single problem, their plans are merged so that the plane moves to each city only once. Plan merging through increasing the justification for replay accomplishes the retracting out of redundant action sequences, which may cause a planning failure. It thus deals with the action-merging interactions defined in (Yang, Nau, & Hendler, 1992). In the next section we describe an empirical study testing the effectiveness of this merging strategy.

### 3.3.1 AN EMPIRICAL TEST OF DERSNLP+EBL'S PLAN MERGING STRATEGY

A preliminary study was conducted to test the effectiveness of DERSNLP+EBL's method of plan merging through replay. This experiment compared DERSNLP+EBL both with and without increasing the justification for replay. The experimental setup consisted of training DERSNLP+EBL on a set of 20 randomly generated 4-goal training problems, and testing on a different set of 30 4-goal test problems. The initial state of each problem contained 12 locations (6 post offices and 6 airports) and 12 transport devices (6 planes and 6 trucks). In the training phase, the planner solved problems and stored the successful plan derivations in the case library. During the testing phase, the planner retrieved multiple stored plan derivations and used these as guidance in solving the test problems. DERSNLP+EBL was tested on the same 30 problems in both replay and from-scratch modes. Replay was either with (DERSNLP+EBL) or without (DERSNLP+EBL-IJ) increased justification. The results are shown in Table 3.

Although overall performance was poorer, the quality of plans in terms of number of steps improved with DERSNLP+EBL's strategy of increasing the justification for step addition. This result suggests that that DERSNLP+EBL's method of plan merging serves to reduce the redundancy in the plans produced through multi-case replay. Recently, Munoz-Avilla and Weberskirch (1997) have tested this *non-redundant merging* strategy in a process planning domain and have found a similar improvements in plan size. The next section describes an evaluation of the full DERSNLP+EBL system.

## 4. Experimental Evaluation of the Complete System

The experiments reported in this section tested the full DERSNLP+EBL system using the dynamic multi-case storage and retrieval strategy described in Section 3. The aim was to evaluate the replay system in a more complex domain. Our hypothesis was that performance would improve over problem solving experience as more negative interactions are discovered and stored. In addition, we predicted that DERSNLP+EBL's method of storage would result in a low library size and low retrieval costs.

The Logistics Transportation domain (Veloso, 1994) has become somewhat of a benchmark in the CBP literature. A scaled up version was therefore chosen for this purpose. We tested large multi-goal problems drawn from the domain shown in Figure 4 scaled up to first 6 and then 15 cities. This size of domain is unusual in the current literature.

### 4.1 Experimental Setup

The experiment was run in phases, each phase corresponding to an increase in problem size. Thirty test problems of each size were randomly generated. Since it is not possible to obtain a truly random distribution within a nonartificial domain, the following strategy was adopted for problem generation. First, the initial state was constructed by fixing the number objects of each type contained in the domain description. For example, in the first experiment, there were six cities (12 locations within cities), six planes, and six trucks. The initial state of each problem was constructed by first including filter conditions (nonachievable conditions). These defined the layout of the cities. For example, the condition (IS-A AIRPORT AP1) identified AP1 as an airport. The condition (SAME-CITY AP1 PO1) indicated that AP1 and PO1 were located in the same city. Second, the achievable (non-filter) conditions that are present in the add clauses of the domain operators were varied for each problem by choosing object constants randomly from those available with the restriction that no two initial state conditions were inconsistent. For example, each plane and package was assigned to a single randomly-chosen location. Goals were chosen from among these achievable conditions in the same manner. Although no attempt was made to create interacting goals, goal interaction was common in the multi-goal problems. This was because a limit was imposed on the number of steps in the plan. It meant that multi-goal problems often could not be solved by concatenating subplans for individual subgoals. In these instances, the planner could take advantage of linking opportunities and achieve multiple goals through common steps. It also meant that often the planner had to backtrack over a derivation for one goal in order to solve an additional goal.

The first experiment used the 6-city domain and was run in 6 phases. The size of the test problems (which ranged from 1 to 6 goals) was increased for each phase. Prior to each phase $n$ of the experiment, the case library was emptied and the planner was retrained on randomly generated problems of size $n$. Training problems were solved by attempting single-goal subproblems from scratch, storing a trace of the derivation of the solution to the problem if one was not already present in the library, and then successively adding an extra goal. Multi-goal problems were stored only when retrieved cases used in solving the problem failed. Whenever a problem could not be solved through sequenced replay of previous cases, the negatively interacting goals contained in the failure reason were identified and a new case achieving these goals alone was stored in the library. In each phase of the experiment, the planner was tested on the same 30 randomly generated test problems after varying amounts of training. The problems were solved both in from-scratch mode and with replay of multiple cases retrieved from the library which had been constructed during training.

| | Logistics (*Best-first CPU limit: 500sec*) | | | | | | |
|---|---|---|---|---|---|---|---|
| Phase | 0 | 20 | 40 | 60 | 80 | 100 | 120 |
| **One Goal** | | | | | | | |
| %Solved | 100%(3) | 100%(3) | 100%(3) | 100%(3) | 100%(3) | 100%(3) | 100%(3) |
| time(sec) | 15 | 14(.1) | 13(.1) | 4(.0) | 5(.10) | 3(.13) | 3(.13) |
| **Two Goal** | | | | | | | |
| % Solved | 90%(4) | 93%(4) | 100%(5) | 100%(5) | 100%(5) | 100%(5) | 100%(5) |
| time(sec) | 1548 | 1069(.2) | 22(1.0) | 23(.2) | 25(.28) | 15(.28) | 11(.26) |
| **Three Goal** | | | | | | | |
| % Solved | 53%(5) | 87%(7) | 93%(7) | 93%(7) | 93%(7) | 100%(8) | 100%(8) |
| time(sec) | 7038 | 2214(.55) | 1209(.49) | 1203(.54) | 1222(.52) | 250(.54) | 134(.58) |
| **Four Goal** | | | | | | | |
| % Solved | 43%(5) | 100%(8) | 100%(8) | 100%(8) | 100%(9) | 100%(9) | 100%(9) |
| time(sec) | 8525 | 563(.99) | 395(.79) | 452(.91) | 24(.97) | 22(.89) | 22(.88) |
| **Five Goal** | | | | | | | |
| % Solved | 0% | 70%(11) | 90%(11) | 93%(11) | 93%(11) | 93%(11) | 100%(12) |
| time(sec) | 15000 | 5269(2) | 2450(1) | 1425(2) | 1479(1) | 1501(1) | 375(1) |
| **Six Goal** | | | | | | | |
| % Solved | 0% | 50%(12) | 70%(13) | 87%(14) | 93%(14) | 93%(14) | 93%(14) |
| time(sec) | 15000 | 7748(3) | 4578(5) | 2191(5) | 1299(3) | 1319(3) | 1244(3) |

Table 4: Performance statistics in Logistics Transportation Domain. Average solution length is shown in parentheses next to %Solved. Case retrieval time is shown in parentheses next to CPU time.

A second experiment which tested the planner on a more complex 15 city domain employed a stable case library formed when DERSNLP+EBL was trained on 120 (6 city, 6 goal) logistics transportation problems. This library of smaller problems was then used when the planner was tested on the larger (15 city) problems ranging from 6 to 10 goals.

## 4.2 Experimental Results

In the first experiment on the 6 city domain DERSNLP+EBL showed substantial improvements with multi-case replay as evident from the results in Table 4. Moreover, replay performance improved with problem-solving experience. The plans that were produced showed only a slight increase in number of steps over the solutions which were obtained in from-scratch mode. The same results are plotted in Figure 17 which graphs cumulative CPU time on all test problems over the six experiments. This figure illustrates how CPU time decreased with the number of training problems solved. The insert shows total CPU time (including case retrieval time) for all of the test problems in the six experiments. As evident in this insert, planning performance improves with increased experience on random problems. However, relatively little experience (20 problems solved) was enough to show significant performance improvements.

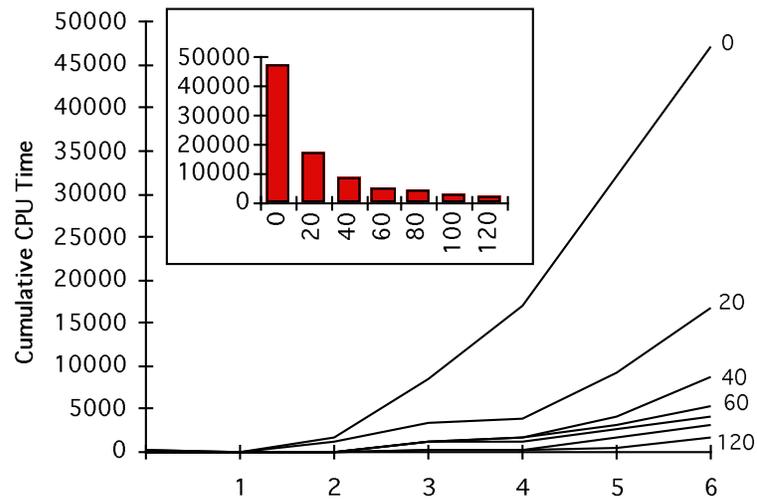

Figure 17: Replay performance in the Logistics Transportation Domain with increasing amounts of training. Thirty problems were tested for each problem size (1 to 6 goals). The amount of time needed to solve all test problems up to that size (including case retrieval time) is shown when problems were solved from scratch (level 0) and with replay after increasing levels of training (after solving 20 ... 120 randomly generated problems). The insert shows the amount of time taken to solve all test problems after increasing amounts of training. A time limit of 500 seconds was placed on problem solving.

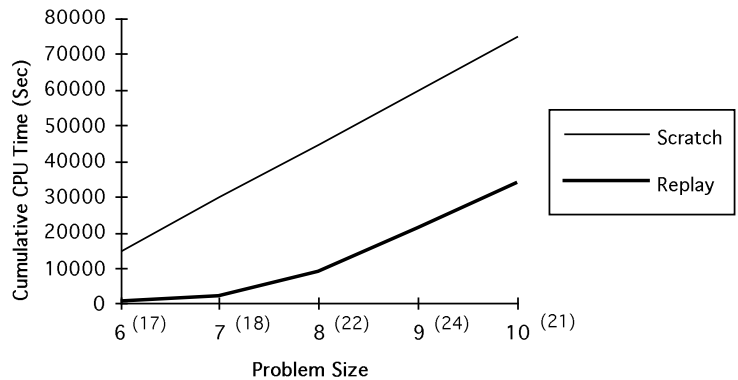

Figure 18: Replay performance in the Logistics Transportation Domain scaled up to 15 cities. A case library was formed as 120 training problems (6 cities, 6 goals) were solved. This library was then used in solving test sets containing larger problems (15 cities, 6 to 10 goals). None of the problems were solved within the time limit (500 sec) in from-scratch mode. For replay mode, average solution length is shown in parentheses next to problem size.

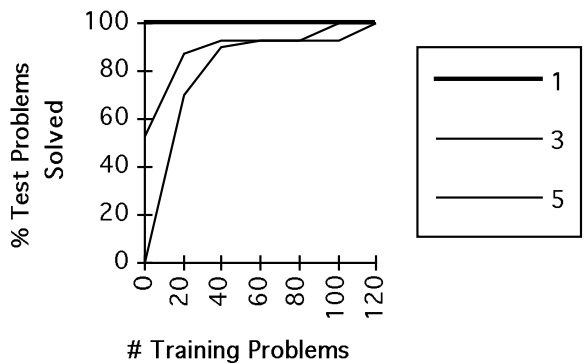

Figure 19: Replay performance in the logistics transportation. The percentage of test problems solved within the time limit (500 sec) is plotted against number of training problems solved. Percentage solved is shown for problems of increasing size (1, 3, and 5 goals).

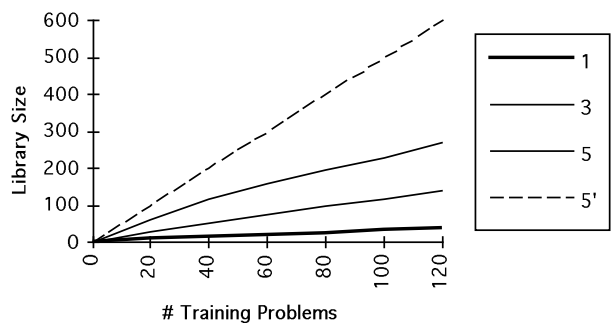

Figure 20: Figure shows the size of the case library with increased number of training problems solved. Library size increases with training problem size (1, 3, and 5 goals). 5' shows the number of single-goal subproblems contained in the 5-goal training problems.

Replay raised the problem-solving horizon, as illustrated in Figure 19. It is more effective with larger problem size, when from-scratch planning tends to exceed the time limit imposed on problem-solving. Figure 20 shows the increase in the size of the library with increasing amounts of training. This figure also indicates that library size is determined more by the amount of interaction in the domain, as opposed to the number of training problems solved. The rate at which the case library grows tapers off and is higher when the planner is trained on larger problems[7].

In the second experiment, a library formed over the course of training on 6-goal problems was used to solve larger problems (6 to 10 goals) in a more complex domain (15 cities) (See Figure 18). None of the larger problems were solved in from-scratch mode within the time limit of 500 sec [8]. The planner continued to maximum time on all problems, indicated in the figure by the linear increase in CPU time. Its performance was substantially better with replay, however. Since library size was relatively small, the improvements in planning performance more than offset the cost of retrieving and adapting previous cases. This finding suggests that the replay strategy employed in these experiments represents an effective method for improving planning performance in complex domains.

---

7. There is more opportunity for interaction in larger problems. For example, a 6-goal problem could contain 6 goals that mutually interact, whereas a 5-goal problem has a maximum of 5 interacting goals.

8. DERSNLP+EBL in from-scratch mode used a best-first strategy. In replay, this best-first strategy is biased so that the subtree under the replayed path is explored first, before the siblings of this path.

| action | (Put-On ?X ?Y ?Z) | action | (New-Tower ?X ?Z) |
|--------|-------------------|--------|-------------------|
| precond | (On ?X ?Z) | precond | (On ?X ?Z) |
| | (Clear ?X) | | (Clear ?X) |
| | (Clear ?Y) | | |
| add | (On ?X ?Y) | add | (On ?X Table) |
| | (Clear ?Z) | | (Clear ?Z) |
| delete | (On ?X ?Z) | delete | (on ?X ?Z) |
| | (Clear ?Y) | | |

Figure 21: The specification of the Blocks World Domain adapted for our experiments.

## 4.3 An Empirical Comparison of DERSNLP+EBL with Rule-Based EBL

Case-based planning and explanation-based learning offer two differing approaches to improving the performance of a planner. Prior research (Kambhampati, 1992) has analyzed their tradeoffs. The hybrid learning approach of DERSNLP+EBL is designed to alleviate the drawbacks associated with both pure case-based planning, and rule-based EBL. Prior to this work, EBL has been used to construct generalized search control rules which may be applied to each new problem-solving situation. These rules are matched at each choice point in the search process (DeJong & Mooney, 1986; Minton, 1990b; Mostow & Bhatnagar, 1987; Kambhampati et al., 1996b). This approach is known to exhibit a *utility problem* since the rule base grows rapidly with increasing problem-solving experience and even a small number of rules may result in a high total match cost (Minton, 1990b; Tambe, Newell, & Rosenbloom, 1990; Kambhampati, 1992; Francis & Ram, 1995). In contrast, the empirical results discussed here (see Table 4) indicate that DERSNLP+EBL has a low case retrieval and match cost.

To demonstrate how DERSNLP+EBL reduces match cost, we conducted an empirical study which compared its performance with UCPOP+EBL, a rule-based search control learning framework (Kambhampati et al., 1996b). This framework constructs reasons for planning failures in a manner similar to DERSNLP+EBL. However, its approach is similar to that of Minton (1990b) in that it employs these explanations in the construction of search control rules which are matched at each node in the search tree. The planners were tested on a set of problems ranging from 2 to 6 goals which were randomly generated from the blocks domain shown in Figure 21. Testing was performed on the same set of thirty problems after increasing amounts of training.

As illustrated in Figure 22, DERSNLP+EBL improved its performance after only 10 training problems solved. UCPOP+EBL failed to improve significantly. The reason is evident in UCPOP+EBL's match time (UCPOP-match) also graphed in Figure 22. For UCPOP+EBL, time spent in matching rules increases with training, wiping out any improvements that may have been gained through the use of those rules. When rules are matched at each choice point in the search tree, a small number of rules is sufficient to substantially increase the total match cost.

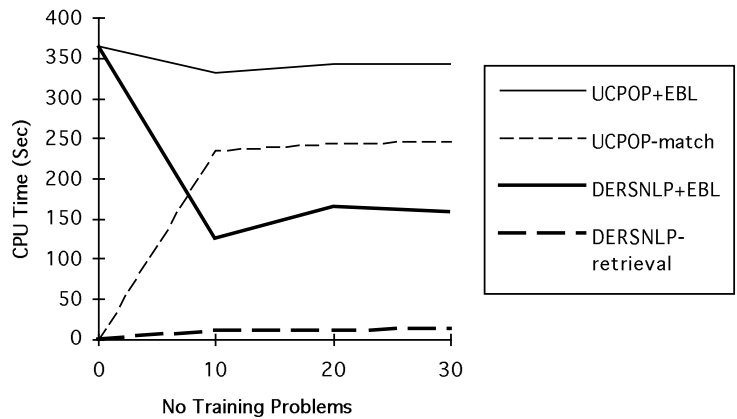

Figure 22: Total CPU Time on 30 blocks world problems after increased amounts of training.

It is also possible to improve the performance of rule-based EBL by reducing the number of rules through the use of utility monitoring strategies (Gratch & DeJong, 1992), or by using a more sophisticated match algorithm (Doorenbos, 1995). For example, Doorenbos (1995) employs an improved rule matcher based on the Rete algorithm. DERSNLP+EBL, on the other hand, aims at alleviating the utility problem by reducing the number of times rules are matched. Similar to rule-based EBL, its learning component is employed to generate rules. However, the rules that are generated govern the retrieval of the cases stored in the library. These are compiled into its indexing structure. DERSNLP+EBL exhibits low match cost by applying retrieval rules at only one point in the search process. Specifically, it retrieves cases only at the start of problem-solving. Each case represents a sequence of choices (a derivation path) thus providing global control as opposed to local. The results shown in Table 4 indicate that the cost of retrieving cases is significantly lower in comparison to time spent in problem-solving.

## 5. Related Work and Discussion

DERSNLP+EBL's storage strategy relies on the capability of the case-based planner to replay multiple cases, each covering a small subset of goals, and then add step orderings to interleave their respective plans. This strategy differs from earlier approaches such as PRIAR (Kambhampati & Hendler, 1992), PRODIGY/ANALOGY (Veloso, 1994), PARIS (Bergmann & Wilke, 1995), and CAPLAN (Munoz-Avila & Weberskirch, 1996), in that the division into goal subsets is not based on the structure of the final plan alone, but on the sequence of events making up the problem-solving episode. Retrieval failures are treated as an opportunity by which the planner stores a new repairing case. In this aspect it is similar to Hammond's CHEF (Hammond, 1990) which also learns to improve its retrieval strategy based on failures. Despite this surface similarity, there are important differences in our approach. DERSNLP+EBL learns from case extension failures, whereas CHEF concentrates on learning from execution failures. Specifically, CHEF assumes an incomplete domain model, consisting of stored cases, and a domain-specific modification theory of *patches*. Given a new problem, CHEF retrieves a previous case, and modifies the retrieved plan using domain specific modification rules to generate a candidate solution for the current problem. The correctness of this solution is then tested with respect to an external causal simulator of the domain. If the solution is found to be incorrect, the explanation of incorrectness (supplied by the simulator) is used to modify the case-library to censor the retrieval of the case in similar situations in the future. This in effect improves the *correctness* of CHEF's domain theory. In contrast, DERSNLP+EBL assumes complete knowledge of the domain, in the form of domain operators. It also has access to a sound and complete plan synthesis strategy. The aim of case-based reasoning in DERSNLP+EBL is to improve the performance of the base-level planner. To this end, DERSNLP+EBL analyzes case extension failures to predict when a case cannot be extended to solve a new problem.

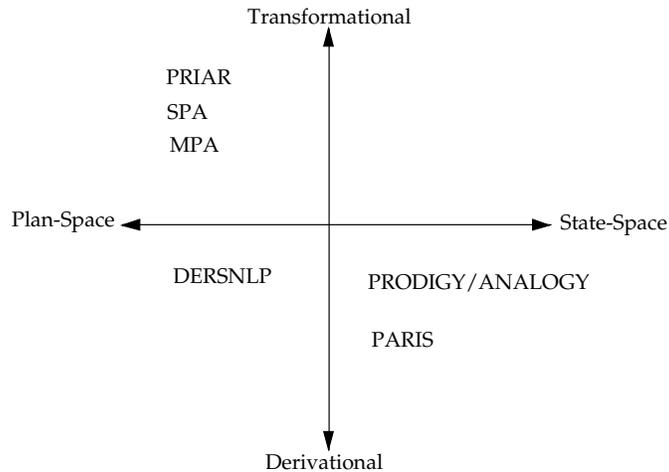

Figure 23: Some different approaches to case-based planning where case adaptation is accomplished by an underlying generative planner.

Fox and Leake (1995) have taken an approach similar to that of CHEF, but use introspective reasoning to explain failures and find repairing cases. Similar to CHEF, introspective reasoning is used to revise indexing in the case library (Fox & Leake, 1995; Ram & Cox, 1994). Other approaches employ domain-specific techniques to improve storage and retrieval from a case library (Munoz-Avila & Weberskirch, 1996; Smyth & Keane, 1995). DERSNLP+EBL differs in that it automatically generates new indices through a well defined and domain-independent methodology (Kambhampati et al., 1996b) which is incorporated into the underlying planning strategy.

Since EBL is employed in explaining case failure as well as success, DERSNLP+EBL complements and extends earlier approaches to case retrieval (Barletta & Mark, 1988; Kambhampati & Hendler, 1992; Hendler, Stoffel, & Mulvehill, 1996; Veloso, 1994; Bergmann & Wilke, 1995; Munoz-Avila & Weberskirch, 1996; Ram & Francis, 1996). Although it exhibits low retrieval and match cost, as with any CBP system, this efficiency may degrade with larger domain size. DERSNLP+EBL's approach is compatible with others aimed at improving match cost (Doorenbos, 1995; Ram & Francis, 1996; Hendler et al., 1996). For example, MPA (Ram & Francis, 1996) is built around a retrieval engine which performs asynchronous memory retrieval. CAPER (Hendler et al., 1996) uses a *structure matching* algorithm which parallelizes the process by which the plan's success conditions represented as a retrieval probe are matched with a large knowledge base of world facts. This process expands binary predicates which match the success conditions into a larger structure containing implicitly specified relations in the knowledge base. This structure acts as a filter, eliminating matches which fail to *line up* with the probe.

DERSNLP+EBL is similar to case-based systems which employ a complete and correct domain-independent planner to generate cases to be stored (Hanks & Weld, 1995; Kambhampati & Hendler, 1992; Koehler, 1994; Veloso, 1994; Ram & Francis, 1996). In surveying this literature, it is possible to distinguish these approaches on two orthogonal scales as shown in Figure 23. In the horizontal direction, the CBP frameworks are ranked as to how the underlying planning strategy falls on a continuum whose end extremes represent the state-space vs plan-space dichotomy. Towards the state-space end of the spectrum is PRODIGY/ANALOGY, which employs the means-ends analysis (MEA) planner, NOLIMIT, to extend a previous case. NOLIMIT is here classed as a state-space planner since it applies actions to the plan based on the current world state and thereby advances the world state.

The PRIAR framework (Kambhampati & Hendler, 1992; Kambhampati, 1994) is based within NONLIN (Tate, 1977). NONLIN creates its plans through hierarchical task reduction. It is also a partial-order (plan-space) planner which constructs plans by protecting their underlying causal structure. Like DERSNLP+EBL, it extends a case through the normal course of plan refinement defined by an underlying plan-space strategy. However, DERSNLP+EBL is implemented within the partial-order, causal-link planner, SNLP (McAllester & Rosenblitt, 1991; Barrett & Weld, 1994). In this aspect it is similar to the SPA system developed by Hanks and Weld (1995).

The different CBP systems may also be distinguished according to their case adaptation strategy. These can be roughly categorized as either *transformational* or *derivational* (Carbonell, 1983; Veloso & Carbonell, 1993b), according to whether they transform a previous plan or replay a previous plan derivation. In the transformational strategies of PRIAR and SPA, the final plan which is the product of the planning episode is stored in the case library. When a case is retrieved this plan is fitted to adapt to the new problem-solving situation by retracting the irrelevant or redundant subparts. Early CBP systems (Carbonell, 1983; Hammond, 1990) also employ transformational techniques to adapt a previous solution. Causal-link planners such as SNLP are ready-made for plan reuse since the causal structure which is employed in plan adaptation is a part of the plan itself. PRIAR and SPA use the plan's causal structure both in fitting the plan to the new problem context, and in extending the fitted plan to solve the new problem. PRIAR differs from SPA in that it employs an *extension-first* strategy. The skeletal plan is first refined through the addition of plan constraints before undertaking any further retraction of constraints. SPA, on the other hand, alternates the retraction of the plan constraints with the further addition of new constraints. MPA (Ram & Francis, 1996) extends SPA's transformational strategy to accomplish multi-case retrieval and adaptation.

As mentioned earlier, derivational analogy is a case-based planning technique which was introduced by Carbonell (Veloso & Carbonell, 1993b). This model was developed by Veloso in PRODIGY/ANALOGY (Veloso, 1994), which employed the case fitting strategy called *derivational replay*. Case fitting based on replay is similar to fitting in plan reuse, in that it is based on the plan's underlying causal structure. The justification for each planning decision which is stored in the derivation trace reflects the causal dependencies between plan steps. Only justified choices are *replayed* in solving the new problem. Replay thus serves the same purpose as retraction in plan reuse. Replay may have an advantage in multi-case reuse since it allows the planner to readily merge small subplans to solve large problems.

DERSNLP can be contrasted to PRODIGY/ANALOGY in that it employs a case fitting methodology called *eager derivation replay* (Ihrig & Kambhampati, 1994a, 1996). With this replay strategy, the applicable cases are replayed in sequence before returning to from-scratch planning. Eager replay simplifies the replay process by avoiding the decision as to how to alternate replay of multiple cases. The effectiveness of this approach is dependent on the underlying plan-space planning strategy (Ihrig & Kambhampati, 1994a). DERSNLP's eager case adaptation strategy allows case failure to be defined in terms of the failure of a single node in the search tree. In particular, case failure is defined as the failure of the *skeletal plan*, which contains all of the constraints that have been adopted on the advice of the previous cases. Eager case adaptation means that explanations of case failure may be constructed through the use of EBL techniques which have been developed to explain analytical failures occurring in the planner's search space.

## 6. Summary and Conclusion

In this paper we have described the design and implementation of the case-based planner, DERSNLP+EBL. The DERSNLP+EBL framework represents an integration of eager case adaptation with failure-based EBL. EBL techniques are employed in building the case library on the basis of experienced retrieval failures. This approach improves on earlier treatments of case retrieval (Barletta & Mark, 1988; Kambhampati & Hendler, 1992; Ihrig & Kambhampati, 1994a; Veloso & Carbonell, 1993a). As a partial-order case-based planner, DERSNLP has the ability to solve large problems by retrieving multiple instances of smaller subproblems and merging these cases through sequenced replay (Ihrig & Kambhampati, 1994a). The DERSNLP+EBL framework extends this approach through the use of new EBL techniques which are employed in the construction of the case library. These techniques are used to explain a plan merging failure and to identify a set of negatively interacting goals. The library is then augmented with a new repairing case covering these interacting goals.

DERSNLP+EBL's method of storing multi-goal cases only when goals are negatively interacting results in a small library size and low retrieval costs. However, multi-case adaptation also involves a tradeoff since effort is expended in merging multiple instances of stored cases. DERSNLP+EBL accomplishes this merging by increasing the justification for replay of step addition decisions. This strategy avoids the addition of redundant steps when goals positively interact. DERSNLP+EBL is therefore aimed at domains such as the Logistics Transportation domain where there is a significant amount positive interaction. It is also aimed at domains where there is negative interaction. It is of course futile to spend effort in explaining case failure if none are encountered.

Section 4 describes an evaluation of the overall efficiency of this storage and retrieval strategy when solving large problems in a complex domain. DERSNLP+EBL shows an improvement in planning performance which more than offsets the added cost entailed in retrieving on failure conditions. The amount of improvement provided by replay shown in these experiments should be seen as a lower bound since a random problem distribution may mean less problem similarity than is found in real world problems.

In conclusion, this paper has described a novel approach to integrating explanation-based learning techniques into case-based planning. This approach has been aimed at issues associated with both pure case-based planning, and with rule-based EBL. In particular, it addresses the mis-retrieval problem of CBP, as well as the utility problem. The results demonstrate that eager case adaptation when combined with DERSNLP+EBL's dynamic case retrieval is an effective method of improving planning performance.

## Acknowledgements

The authors wish to thank Amol D. Mali, Eric Lambrecht, Eric Parker, and the anonymous reviewers for their helpful comments on earlier versions of this paper. Thanks are due to Terry Zimmerman for providing insight into UCPOP+EBL. This research is supported in part by NSF Research Initiation Award IRI-9210997, NSF Young Investigator award IRI-9457634, and the ARPA Planning Initiative grants F30602093-C-0039 (phase II) and F30602-95-C-0247 (phase III).